\pdfoutput=1

\documentclass[11pt]{article}

\usepackage{acl}

\usepackage{times}
\usepackage{latexsym}
\usepackage{multirow}
\usepackage{graphicx} 

\usepackage{booktabs}
\usepackage{algorithm}
\usepackage{wrapfig}
\usepackage{lipsum}
\usepackage{amsmath}
\usepackage{linguex}

\usepackage[T1]{fontenc}

\usepackage[utf8]{inputenc}

\usepackage{microtype}

\usepackage{inconsolata}

\usepackage{graphicx}

\title{Do LLMs Recognize \textit{me}, When \textit{I} is not \textit{me}:\\
Assessment of LLMs Understanding of Turkish Indexical Pronouns in Indexical Shift Contexts}

\author{Metehan Oğuz $^*$  \\
  USC\\ 
  \texttt{moguz@usc.edu} \\\And
  Yusuf Umut Ciftci$^*$ \\
  USC \\
  \texttt{yciftci@usc.edu} \\\And
  Yavuz Faruk Bakman$^*$ \\
  USC \\
  \texttt{ybakman@usc.edu} \\}

\begin{document}
\maketitle

\def\thefootnote{*}\footnotetext{Equal Contribution}\def\thefootnote{\arabic{footnote}}
\begin{abstract}

Large language models (LLMs) have shown impressive capabilities in tasks such as machine translation, text summarization, question answering, and solving complex mathematical problems. However, their primary training on data-rich languages like English limits their performance in low-resource languages. This study addresses this gap by focusing on the Indexical Shift problem in Turkish. The Indexical Shift problem involves resolving pronouns in indexical shift contexts, a grammatical challenge not present in high-resource languages like English. We present the first study examining indexical shift in any language, releasing a Turkish dataset specifically designed for this purpose. Our Indexical Shift Dataset consists of 156 multiple-choice questions, each annotated with necessary linguistic details, to evaluate LLMs in a few-shot setting. We evaluate recent multilingual LLMs, including GPT-4, GPT-3.5, Cohere-AYA, Trendyol-LLM, and Turkcell-LLM, using this dataset. Our analysis reveals that even advanced models like GPT-4 struggle with the grammatical nuances of indexical shift in Turkish, achieving only moderate performance. These findings underscore the need for focused research on the grammatical challenges posed by low-resource languages. We released the dataset and code \href{https://anonymous.4open.science/r/indexical_shift_llm-E1B4} {here}.  
\end{abstract}

\section{Introduction}

Large language models demonstrate remarkable capabilities in zero-shot and few-shot learning, excelling across a diverse range of tasks such as machine translation, text summarization, question answering, and solving complex mathematical problems \cite{ye2023comprehensive, openai2024gpt4, touvron2023llama2}. However, most large language models (LLMs) are primarily trained on data-rich languages like English, and their performance evaluations are also conducted in these languages \cite{2024aya}. Consequently, this focus on data-rich languages may lead to the under-exploration of challenges unique to low-resource languages.

Recent studies have evaluated the performance of large language models on linguistic tasks such as coreference resolution to examine their ability to match expressions referring to the same entity \cite{assessllmcoreference, arellmscoreference, LLMsfewshot, gpt2pronoun_res, coreference_clinical}.
In this study, we investigate LLMs' performance on interpreting indexical pronouns, with a focus on the Indexical Shift problem, a unique linguistic challenge related to but distinct from pronoun resolution, primarily encountered in low-resource languages like Turkish \citep{sener-sener:2011}, Amharic \citep{schlenker:1999}, Zazaki \citep{anand:2004}, Uyghur \citep{shklovsky:2014}, Nez Perce \citep{deal:2020} and Japanese \citep{sudo:2012}.

Indexical elements like \textit{I} and \textit{here} refer to the referents of the speech context such as the speaker or location of utterance. In most languages, these elements must be interpreted within the actual speech context, referring to the actual speaker or location of utterance. However, indexical shift occurs in some languages, like Turkish, where an indexical element can refer to the referents of the reported context, rather than the actual speech context (see Section \ref{indexical} for details). 

Indexical elements are substantially different from pronouns regarding what antecedents they can refer to and what factors restrict their interpretations. For example, while pronouns are almost always ambiguous and can refer to a wide range of entities, first person indexical unambiguously refers to the speaker of the utterance. Even in languages that allow indexical shift, the first person indexical is ambiguous between only two possible referents (the speaker vs the attitude holder), being interpreted based on contex/world knowledge. Moreover, Turkish allows indexical shift only in some syntactic structures (finite embedded clauses) but not in others (e.g. nominalized embedded clauses), which makes indexical shift a unique challenge, requiring attention to specific syntactic rules and context (see Section \ref{related} for a detailed comparison of indexical elements and pronouns regarding coreference resolution).





We investigate the capability of multilingual large language models to handle pronoun resolution within the context of indexical shift in Turkish. To the best of our knowledge, this is the first study examining indexical shift in any language. Therefore, we have released a Turkish dataset specifically designed to evaluate LLMs on the indexical shift problem. Our contributions in this work are as follows: 
\begin{itemize}
    \item We released the Indexical Shift Dataset in Turkish, comprising 156 multiple-choice questions to evaluate LLMs on the indexical shift problem in few-shot setting. Each sample in this dataset includes the necessary linguistic details.
    \item We evaluate recent multilingual LLMs, including GPT-4, GPT-3.5 \cite{openai2024gpt4}, Cohere-AYA \cite{2024aya}, Trendyol-LLM \cite{trendyol_llm_7b_base_v01}, and Turkcell-LLM \cite{turkcell_llm_7b_v1}, using our dataset. We statistically analyze the factors that influence these models' decisions.
    \item We conclude that even advanced models like GPT-4 struggle to grasp the grammatical nuances of indexical shift in Turkish, showing only moderate performance at best. These findings highlight the need for a special focus on the grammatical challenges of low-resource languages.
\end{itemize}

\section{Indexical Shift in Turkish}\label{indexical}
\paragraph{Indexical elements.} Indexical elements such as English \textit{I, you, here} and \textit{yesterday} are used to refer to referents of the speech-act coordinates \citep{kaplan:1977, schlenker:2003, anand:2004, deal:2020}. For example, \textit{I} is used to refer to \textit{author} (speaker) of the utterance, while \textit{here} is used to refer to the \textit{location} where the utterance was made, and thus sentences like \ref{Eng} mean different things if uttered by different people and/or in different locations. If \ref{Eng1} is uttered by \textit{John} in \textit{Los Angeles}, it means that John was born in Los Angeles, but if it is uttered by \textit{Mary} in \textit{Boston} it means that Mary was born in Boston. 

\ex. \label{Eng}
\a. \textbf{I} was born \textbf{here}. \label{Eng1} 
\b. Peter thinks that \textbf{I} went to Atlanta.\label{Eng2} 


\noindent In most languages, including English, indexical elements must always be interpreted inside the actual speech context, referring to actual speech-act coordinates (e.g. author, location).\footnote{One exception for this generalization is direct quotation (e.g. Peter said/thought, `I went to Atlanta'.), where quoted material is interpreted as verbatim utterance/thought produced by its owner. Direct quotation is out of the scope of this paper.} So, if \ref{Eng2} is uttered by \textit{John}, the indexical \textit{I} can only be interpreted as referring to John (e.g. John is believed to have gone to Atlanta) but nobody else. Importantly, even though Peter's beliefs are reported in \ref{Eng2}, the indexical element \textit{I} cannot be interpreted as referring to Peter (author of the reported belief), but must be interpreted as referring to John (author of the actual sentence).

\paragraph{Indexical Shift.} Turkish allows indexical shift \citep[e.g.][]{sener-sener:2011}, a situation where an indexical element gets its referent from the reported context, rather than the actual context of utterance. For instance, the Turkish first person indexical \textit{ben} in \ref{Turk1} can refer to the attitude holder \textit{Burak}, who is the author of the reported belief, or to the author/speaker of the actual sentence.\footnote{Turkish is a \textit{pro}-drop language, meaning that the subject of the clause can be dropped (phonologically null). In this example, and henceforth, parentheses indicate that the subject can optionally be dropped. When subject is dropped, its person features are indicated by the agreement marker on the verb.}

\ex.
\gll Burak yine [ (\textbf{ben}) mezun ol-du-m ] san-ıyor. \\
	Burak again \phantom{s} \textsc{1sg} graduate be-\textsc{pst-1sg} \phantom{s} think-\textsc{prog}\\
\glt `Burak thinks again that \{he/\textit{speaker}\} graduated.' \label{Turk1}


\noindent In this regard, sentences like \ref{Turk1} are ambiguous between readings where first person indexical \textit{ben} is shifted (referring to the attitude holder \textit{Burak}) or non-shifted (referring to actual speaker), and Turkish speakers interpret such sentences based on previous context or upcoming sentences \citep{kuram:2020}. For example, in a context where the actual speaker is the conversation topic, \ref{Turk1} would naturally be interpreted in the non-shifted reading (I = speaker), but if the conversation is about \textit{Burak}, the sentence would be naturally interpreted in the shifted reading (I = Burak).\footnote{Some readers may wonder if embedded material in sentences like \ref{Turk1} is direct quotation (e.g. \textit{Peter thinks, `I am smart.'}, in English). \citet{ozyildiz:2012} and \citet{oguz-etal:2020} use linguistic diagnostics to show that these are not instances of direct quotation but are true instances of indexical shift.}

\begin{table*}[]
\resizebox{\textwidth}{!}{%
\begin{tabular}{@{}ccccc@{}}
\multicolumn{4}{c}{\multirow{2}{*}{}} &
  \textbf{Question} \\
\multicolumn{4}{c}{} &
  NAMENull kimin Almanca bildiğini sanıyor? \\ \midrule
\textbf{Context} &
  \textbf{Context prime} &
  \textbf{Sentence type} &
  \textbf{Sentence} &
  \textbf{Ground truth} \\ \midrule
\multirow{2}{*}{\begin{tabular}[c]{@{}c@{}}Merhaba, ben SPEAKER. Ankara'da yaşayan bir öğrenciyim.\\ NAMENull diye bir arkadaşım var. On tane Almanca kelime öğrenmiş.\\ “Hi, my name is SPEAKER. I am a student living in Ankara. \\ I have a friend named NAMENull. He learned ten German words.”\end{tabular}} &
  \multirow{2}{*}{Shifted} &
  Finite &
  \begin{tabular}[c]{@{}c@{}}NAMENull Almanca biliyorum sanıyor.\\ “NAMEnull thinks he knows German.”\end{tabular} &
  Shifted \\ \cmidrule(l){3-5} 
 &
   &
  Nominalized &
  \begin{tabular}[c]{@{}c@{}}NAMENull Almanca bildiğimi sanıyor.\\ “NAMEnull thinks I know German.”\end{tabular} &
  Speaker \\ \midrule
\multirow{2}{*}{\begin{tabular}[c]{@{}c@{}}Merhaba, ben SPEAKER. Ankara'da yaşayan bir öğrenciyim. NAMENull diye bir arkadaşım var. \\ NAMENull söylediklerini havaalanındaki turistler için Almanca'ya çevirmemi istedi.\\ “Hello, I'm SPEAKER. I am a student living in Ankara. I have a friend named NAMEnull. \\ NAMEnull asked me to translate what he said into German for the tourists at the airport.”\end{tabular}} &
  \multirow{2}{*}{Speaker} &
  Finite &
  \begin{tabular}[c]{@{}c@{}}NAMENull Almanca biliyorum sanıyor.\\ “NAMEnull thinks I know German.”\end{tabular} &
  Speaker \\ \cmidrule(l){3-5} 
 &
   &
  Nominalized &
  \begin{tabular}[c]{@{}c@{}}NAMENull Almanca bildiğimi sanıyor.\\ “NAMEnull thinks I know German.”\end{tabular} &
  Speaker \\ \bottomrule
\end{tabular}%
}
\caption{An example four context-sentence pairs from the dataset.}
\label{tab:dataset_table}
\vskip -0.2in
\end{table*}

\paragraph{Syntactic Restrictions on Indexical Shift.} Indexical shift is a quite rare syntactic/grammatical property, observed in a small set of languages like Amharic \citep{schlenker:1999}, Zazaki \citep{anand:2004}, Uyghur \citep{shklovsky:2014}, Nez Perce \citep{deal:2020}. These languages are different from others (e.g. English) in that their syntax possesses the necessary machinery to allow indexical shift (see \citet{deal:2020} for theoretical details and a comprehensive list of languages that allow indexical shift). Moreover, even within a language, indexical shift might be allowed or disallowed depending on the syntactic structure of a sentence. For example, indexical shift in Turkish is observed only with finite embedded clauses like \ref{Turk1}, but is not allowed in other grammatical structures such as nominalized embedded clauses like \ref{Turk2}, formed by the nominalizer suffix \textit{-DIK} on the embedded verb.

\ex.
\gll Burak yine [ (\textbf{ben}-im) mezun ol-duğ-um-u ] san-ıyor. \\
	Burak again \phantom{s} \textsc{1sg-gen} graduate be-\textsc{nmlz-1sg-acc} \phantom{s} think-\textsc{prog}\\
\glt `Burak thinks again that \{*he/\textit{speaker}\} graduated.' \label{Turk2}


\noindent Different from \ref{Turk1}, the first person indexical \textit{ben} in \ref{Turk2} can only refer to the actual speaker of the sentence (similar to English), regardless of the context it is uttered in (e.g. it cannot undergo indexical shift and refer to the attitude holder Burak). This contrast between \ref{Turk1} and \ref{Turk2} is due to the syntactic/grammatical properties of the finite and nominalized embedded clauses in Turkish, which are acquired by the native speakers of the language (see \citet{sener-sener:2011} and \citet{oguz-etal:2020} for syntactic details).

In this study, we aim to test whether LLMs are able to capture this grammatical contrast between two embedded clause types and successfully interpret indexical elements in syntactic environments that allow (e.g. finite embedded clauses) or block indexical shift (e.g. nominalized embedded clauses).

\section{Related Work}\label{related}




There are substantial differences between pronouns and indexical elements. To begin with, even though syntactic/semantic factors can influence pronoun resolution by making some nouns more likely antecedents of pronouns (e.g. subject bias), they do not totally rule out other nouns as possible referents.\footnote{Except for ones that violate syntactic principles like the Binding Theory \cite{chomsky:1981}.} For example, previous work suggests that speakers mostly interpret the third person pronoun \textit{he} in \ref{Eng3} as referring to the subject \textit{John} (for syntactic or contextual reasons), but the object \textit{Bill} is still a possible antecedent, meaning that the sentence is ambiguous \cite[e.g.][]{crawley-etal:1990, stewart-pickering:1998, pickering-majid:2007}.

\ex. John hit Bill and \textbf{he} ran away. \label{Eng3}

\noindent Moreover, pronouns can refer to nouns that are contextually salient, but not present in the sentence. For example, the third person pronoun \textit{he} can be interpreted as referring to a contextually salient person named \textit{Peter}. As a result, context plays a crucial role in how speakers interpret pronouns. Previous work in the field (cited above) show that LLMs are able to use contextual information during coreference resolution and show good performance. 

Indexical elements, on the other hand, must unambiguously refer to the discourse coordinates (e.g. speaker), except for indexical shifting environments. In syntactic contexts where indexical shift is allowed (e.g. Turkish finite embedded clauses), indexical elements are similar to pronouns in that their referent can be ambiguous. However, indexical elements are still different from pronouns in that they are ambiguous between only two referents (speaker vs attitude holder), while pronouns are technically free to refer to an unlimited amount of antecedents (that can be salient in context). 

In summary, indexical elements are restricted by different syntactic factors than pronouns (e.g. clause type) and are usually unambiguous. Moreover, even in contexts where indexical shift is possible, indexical elements are restricted to two possible referents, depending on whether indexical shift takes place or not, while pronouns are free to refer to a wide range of entities. Thus, indexical elements and indexical shift create a unique challenge for LLMs, requiring to take into account the syntactic constraints regarding indexical shift while also employing general coreference resolution strategies like contextual information.

\section{Turkish Indexical Shift Dataset}







To test LLMs' ability to understand indexical shift in Turkish, we created a dataset containing 156 entries with sentences containing the Turkish first person indexical (silent/dropped) that could potentially refer to the speaker (non-shifted) or to the attitude holder of the clause (shifted). 

Since the interpretation of indexical elements in Turkish (shifted vs non-shifted) depend on the context they appear in, we created two contexts for each experimental sentence: one priming the shifted reading, the other priming the non-shifted reading of the first-person indexical. Moreover, for each context, we created a version of the sentence with a nominalized embedded clause like \ref{Turk2} (rather than finite embedded clause like \ref{Turk1}), where indexical shift is not allowed by the grammar (e.g. the indexical must refer to the speaker even if the reported context priming otherwise). Together, this four context-sentence pairs lead to four different classes as summarized in Table \ref{tab:dataset_table}. 

We used sentences with three different verbs that allow indexical shift in Turkish: \textit{iste} `to want', \textit{san} `to think/believe', \textit{düşün} `to think'. These verbs trigger specific morphosyntactic requirements in Turkish. For example, \textit{iste} `to want' requires a subjunctive marker on the embedded verb, while \textit{san} `to think/believe' and \textit{düşün} `to think' require regular tense morphology. Also, \textit{düşün} `to think' requires the complementizer \textit{diye} while the other two does not require/allow \textit{diye}.

Each entry in the dataset have the following information:
\begin{itemize}
    \item The embedding verb used in the sentence.
    \item The context first person indexical meaning that the context encourages (context prime).
    \item Ground truth entity that the indexical is referring to (SPEAKER or SHIFTED).
    \item The sentence.
    \item Sentence type (Nominalized or Finite).
    \item Question to reveal the LLM's interpretation of the indexical.
\end{itemize}



In order to augment our dataset with different name pairs for testing, SPEAKER and NAMENull placeholders are used instead of the speaker name and the reported third party with NAME-\textsc{acc},  NAME-\textsc{gen}, NAME-\textsc{dat}, NAME-\textsc{loc}, NAME-\textsc{com} were used for the accusative, genitive, dative, locative, and comitative forms of the third person subject's name. 


The Turkish language indexical shift test dataset is open sourced along with the associated source code, \href{https://anonymous.4open.science/r/indexical_shift_llm-E1B4} {here}.


\begin{table*}
\caption{Precision, recall, and f1 performances of each model for each class and their macro averages.}
\centering
\fontsize{10}{12}\selectfont
\begin{tabular}{l ccc cc ccc cc ccc}
\toprule
 & \multicolumn{3}{c}{\textbf{Speaker}} &&& \multicolumn{3}{c}{\textbf{Shifted}} &&& \multicolumn{3}{c}{\textbf{Macro Average}}\\
 \cmidrule{2-4} \cmidrule{7-9} \cmidrule{12-14} 
Model & Precision & Recall & F1 &&& Precision & Recall & F1 &&& Precision & Recall & F1 \\
\midrule 
GPT-4& \textbf{0.91}&0.69&0.78 &&& 0.46&\textbf{0.79}&\textbf{0.58} &&& \textbf{0.68}&\textbf{0.74}&\textbf{0.68}  \\
GPT-3.5& 0.77&0.59&0.67 &&& 0.28&0.47&0.35 &&& 0.53&0.53&0.51  \\
Cohere-AYA& 0.83&\textbf{0.84}&\textbf{0.84} &&& \textbf{0.51}&0.50& 0.51 &&& \textbf{0.67}&0.67&\textbf{0.67}  \\
Trendyol-LLM& 0.71&0.49&0.58 &&& 0.22&0.42&0.29 &&& 0.47&0.45&0.43  \\
Turkcell-LLM& 0.88&0.12&0.22 &&& 0.27&0.95&0.42 &&& 0.57&0.54&0.32  \\
\bottomrule
\end{tabular}
\label{accuracy_table}
\end{table*}

\section{Experiments}
In this section, we explain our experimental setup, results, and discussion of the results.
\subsection{Experimental Design}

\paragraph{Models.} We assess the performance of five language models: GPT-4 \cite{openai2024gpt4}, GPT-3.5 \cite{brown2020language}, Cohere-AYA \cite{2024aya}, Trendyol-LLM \cite{trendyol_llm_7b_base_v01}, and Turkcell-LLM \cite{turkcell_llm_7b_v1}. Both GPT-4 and GPT-3.5 are closed-source, advanced multilingual models. Cohere-AYA, a 13-billion parameter model, is trained in 101 languages and is built by fine-tuning the mT5 model \cite{xue2021mt5}. Trendyol-LLM is based on the LLama-2 7-billion model and fine-tuned on both Turkish and English data. Lastly, Turkcell-LLM is a fine-tuned version of the Mistral 7-billion model, specifically adapted for Turkish data.

\begin{figure*}[!htbp]
\begin{center}
\includegraphics[width=1.0\textwidth]{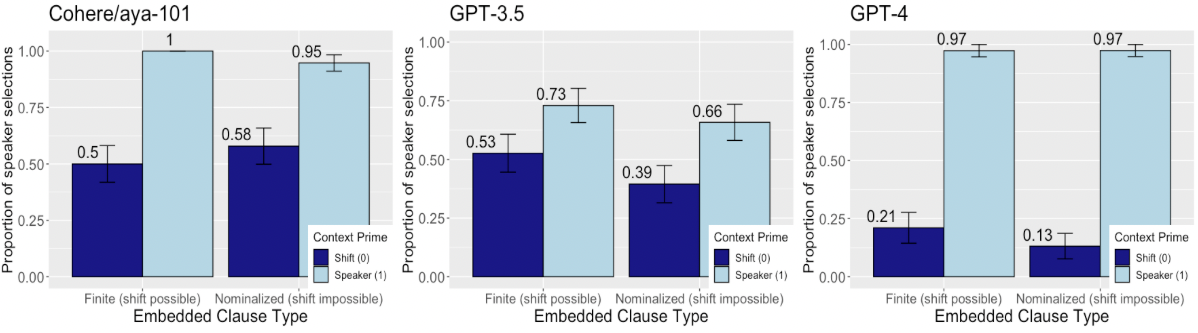}
\caption{Selection analysis plot of GPT-4, GPT-3.5 and Cohere-AYA models. Their outputs are significantly influenced by the context prime, indicating the context's meaning toward either the speaker or the shifted class. No other significant factors were observed.}
\label{selection_main}
\end{center}
\vskip -0.2in
\end{figure*}

\paragraph{Evaluation Strategy.}
We evaluate the performance of models using a multiple-choice question-answer format similar to the Massive Multitask Language Understanding benchmark (MMLU) \cite{hendrycks2021measuring}. The questions are presented in Turkish, following this template:
\begin{verbatim}
{in_context_learning_examples}
Soru: {context} {question}?

Seçenekler:
A. {choice_a}
B. {choice_b}

Doğru cevap:
\end{verbatim}
where \verb|Soru| means "question", \verb|Seçenekler| means "choices", and \verb|Doğru cevap| means "correct answer". We select five random examples from our dataset as in-context examples and evaluate the remainder. For open-source models, we calculate the probabilities of the tokens "A" and "B" to determine the most likely answer. For closed-source models, we modify the prompt by adding: \emph{``Aşağıdaki soruları cevapla. Sadece cevap olarak A veya B yazman lazım.''}\footnote{In English: Answer the following questions. You need to write only A or B as your answer.} to ensure accurate response generation. Both GPT models comply strictly with this rule, generating only the letters A or B as responses.

\begin{table}
\caption{Accuracy results for each clause type: finite and nominalized. All models except Cohere-AYA shows worse performance in nominalized sentences.}
\centering
\fontsize{10}{12}\selectfont
\begin{tabular}{l ccc}
\toprule
Model & Finite & Nominalized & Average\\
\midrule 
GPT-4& \textbf{0.88} & 0.55 & 0.72 \\
GPT-3.5&0.60 & 0.53& 0.57 \\
Cohere-AYA&0.75 & \textbf{0.76}& \textbf{0.76}\\
Trendyol-LLM&0.50&0.50 & 0.50\\
Turkcell-LLM&0.57 & 0.09& 0.33\\
\bottomrule
\end{tabular}
\label{clause_acc}
\end{table}

Our dataset originally contains placeholders such as "NAMENull" and "SPEAKER" instead of real names. We replace these placeholders with actual random Turkish names to provide more natural linguistic contexts. To decrease the effect of potential gender biases within the models, we uniformly use either female or male names for all placeholders within a single question. 


Lastly, to address the choice bias demonstrated in prior studies \cite{zheng2024large}, we implement random assignment of choice options in both the question prompts and in-context examples. Additionally, for open-source models, we enhance reliability by presenting each question twice with the order of choices reversed. We then aggregate the probabilities assigned by the model to each option across these iterations to determine the most likely choice. This method helps to mitigate any inherent preference the model may have towards the position of the answer choices.

\begin{figure}[]
\begin{center}
\includegraphics[width=0.47\textwidth]{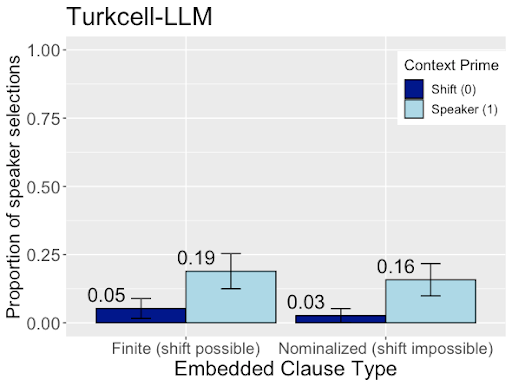}
\caption{Selection analysis plot of Turkcell-LLM. Neither clause type nor context prime has a statistically significant effect.}
\label{turkcell}
\end{center}
\vskip -0.3in
\end{figure}

\paragraph{Metrics.}
Our dataset exhibits a class imbalance, with 75\% of the ground truth labeled as SPEAKER and the remaining 25\% as SHIFTED. Given this imbalance, accuracy alone would not provide a comprehensive measure of model performance; a trivial model that consistently outputs "SPEAKER" could achieve 75\% accuracy without truly understanding the data. To address this, we follow the common practice in the literature \cite{branco2015survey} and report precision, recall, and F1 scores for both the SPEAKER and SHIFTED classes. We also compute the macro precision, recall and F1 score to summarize overall performance. Moreover, to see the performance of the models in different clause types (finite or nominalized), we provide the accuracy of all models in different clause types and average accuracy as well. Lastly, analyze factors influencing an LLM’s decision-making on a given question, using R Software \cite{Rstudio} to build the best fitting Linear Mixed-Effect Regression (LMER) model \citep{bates-etal:2015} with item as the random factor and model selection as the dependent variable. The findings of our statistical analyses are discussed in detail in Section \ref{logistic}.


\subsection{Main Results}
The performance of all models is presented in Table \ref{accuracy_table}. For the Speaker class, GPT-4 achieves the highest precision, while Cohere-AYA consistently delivers high precision and recall, resulting in the highest F1 score for this class. For the Shifted class, GPT-4 attains a maximum F1 score of 0.58, significantly lower than its performance for the Speaker class. All models, except Turkcell-LLM, exhibit lower performance for the Shifted class, indicating a tendency to make mistakes when either the ground truth or the model output is the class Shifted (low precision and low recall). 

Examining the macro average results, we observe that GPT-4 and Cohere-AYA have comparable and highest F1 performance, whereas other models are behind of them with a significant margin. This performance gap for GPT-4 can be attributed to its advanced capabilities, likely due to a large training corpus and model size \cite{openai2024gpt4}. Similarly, the performance gap for Cohere-AYA may be due to its training data, which includes many Turkish samples \cite{2024aya}. However, even the performance of GPT-4 and Cohere-AYA is far from optimal. Lastly, as shown in Table \ref{clause_acc}, GPT-4 and Cohere-AYA perform relatively well at predicting indexical shift in sentences with finite clauses, where shift is possible. However, in sentences with nominalized clauses, where shift is not possible, the performance of all models drops significantly. Among them, only the Cohere-AYA model demonstrates a significantly better prediction accuracy than random guessing (50\%). This finding highlights the need for greater attention to the grammatical challenges in low-resource languages.

\subsection{Which Factors Effect LLMs' Decision?}\label{logistic}

In this section, we employ Linear Mixed-Effect Regression (LMER) models to measure the impact of various factors in the dataset on LLM decisions. These factors include sentence type (\textit{finite} vs \textit{nominalized}), gender, and context prime (priming \textit{shifted} vs \textit{non-shifted} readings). Through this analysis, we observe that LLM behaviors can be clustered based on their responses to our task. Below, we examine each LLM cluster and their behavior patterns in detail.

\paragraph{GPT Family and Cohere-AYA.} The decisions of these three models are influenced by the context prime, indicating that the context's meaning leans towards either the speaker or shifted class. This effect is statistically significant (\textit{p}'s < 0.001). As illustrated in Figure \ref{selection_main}, the models' decisions change significantly when the context prime is altered (represented by dark blue and blue colors). For instance, Cohere-AYA selects the speaker class 100\% of the time when the context prime indicates the speaker in finite sentences, but this proportion drops to 50\% when the context prime indicates the shifted class. This substantial difference in selection proportions highlights the significant impact of the context prime, an effect that is also observed in GPT models across different sentence types. 

Aside from the effect of context prime, all other factors have a statistically non-significant impact on the models' decisions (\textit{p}'s > .05). This is interesting because, for a native speaker, the  clause type (finite or nominalized) directly influences the interpretation of the sentences, allowing indexical shift in finite embedded clauses \ref{Turk1} but not in nominalized embedded clauses \ref{Turk2}.

\begin{figure}[]
\begin{center}
\includegraphics[width=0.47\textwidth]{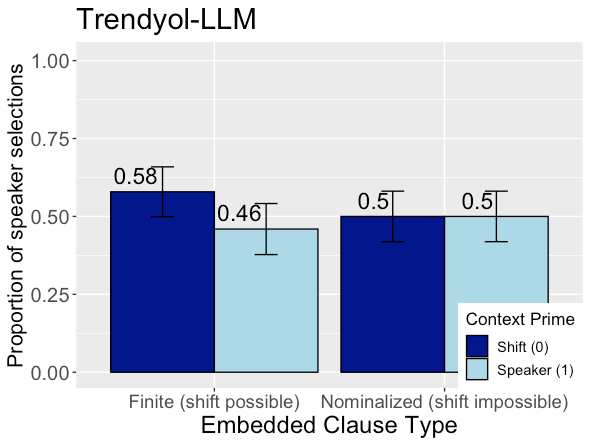}
\caption{Selection analysis plot of Trendyol-LLM. Neither clause type nor context prime has a statistically significant effect.}
\label{trendyol}
\end{center}
\vskip -0.3in
\end{figure}

\paragraph{Trendyol-LLM and Turkcell-LLM.} Figures \ref{turkcell} and \ref{trendyol} illustrate the mean decisions of these models in each item class. Our analyses show that the decisions of these models are not influenced by either context prime or clause type (\textit{p}'s > .05). Specifically, no factors significantly affect the decisions of Turkcell-LLM, while the only factor that affects the decisions of Trendyol-LLM is interestingly gender (\textit{p} < .001). Furthermore, Turkcell-LLM almost consistently outputs the shifted class, as seen in Figure \ref{turkcell}. Given their comparatively low performance, we interpret these results to mean that these two models lack the reasoning capability to understand the indexical shift problem in Turkish and produce reasonable outputs.

\paragraph{Summary.} Overall, none of the models tested in this study are sensitive to clause type, showing that all of these models fail to learn the grammatical grounds where indexical shift is possible (finite embedded clauses) or not (nominalized embedded clauses). Trendyol-LLM and Turkcell-LLM struggle significantly with interpreting the task. The decisions of the other models are primarily affected by the context prime, mimicking native speaker behavior with finite embedded clauses, but overgeneralizing this behavior with nominalized embedded clauses, where context prime does not play a role for native speakers (since indexical shift is not available).

\section{Conclusion}

In this study, we assess large language models (LLMs) on pronoun resolution tasks within indexical shift contexts, focusing specifically on a low-resource language, Turkish. To facilitate this evaluation, we release a Turkish indexical shift dataset comprising 156 samples. We test recent multilingual models on this dataset and find their performance lacking. Additionally, we observe that none of the LLMs' decisions are influenced by grammatical nuances, such as finite versus nominalized clauses, which contrasts with the behavior of native speakers. Our findings highlight the need for greater attention to the grammatical challenges of low-resource languages in the development and evaluation of LLMs.

\section{Limitations}
One limitation of the current study is that it concentrated solely on the first person indexical in Turkish, which was due to linguistic limitations regarding indexical shift in Turkish. As explained in detail by \citet{deal:2020}, indexical elements within a language do not need to show a uniform behavior, and can have different properties than one another. For example, in Turkish, the person indexicals \textit{ben} `I' and \textit{sen} `you' and the temporal indexical \textit{yarın} `tomorrow' allow indexical shift, while the locative indexical \textit{bura} `here' does not allow indexical shift \cite{oguz-etal:2020}. In other words, the locative indexical \textit{bura} `here' cannot undergo indexical shift, and it must always be interpreted as the location where the sentence was uttered (similar to English \textit{here}). Considering this, the locative indexical \textit{bura} `here' could not be included in our study. Moreover, the second person indexical \textit{sen} can only shift under the verb \textit{de} `to say', and cannot shift under other verbs like \textit{san} `to think' or \textit{iste} `to want' (because they cannot take an addressee). In the current study, we aimed to observe LLMs' performance under various types of embedding verbs, while keeping our experimental sentences maximally consistent. For this reason, we could not investigate the second person indexical \textit{sen}, which does not allow indexical shift in any other verb than \textit{de} `to say'. Future work can extend our findings by investigating LLMs' performance with other indexical elements than the first person \textit{ben}.

\bibliography{custom}

\end{document}